\documentclass[letterpaper, 10 pt, conference]{ieeeconf}  % Comment this line out if you need a4paper

\IEEEoverridecommandlockouts                              % This command is only needed if 
\usepackage{graphics} % for pdf, bitmapped graphics files
\graphicspath{ {./images/} }
\usepackage{subcaption}

\usepackage{amsmath} % assumes amsmath package installed
\DeclareMathOperator*{\argmax}{argmax}
\usepackage{amssymb}  % assumes amsmath package installed
\usepackage{algorithm}
\usepackage[noend]{algpseudocode}
\usepackage{lipsum}
\usepackage{mwe}
\usepackage{gensymb}
\usepackage{amsthm}
\usepackage{float}
\usepackage{hyperref}
\newtheoremstyle{mystyle}%                % Name
  {}%                                     % Space above
  {}%                                     % Space below
  {\itshape}%                                     % Body font
  {}%                                     % Indent amount
  {\bfseries}%                            % Theorem head font
  {.}%                                    % Punctuation after theorem head
  { }%                                    % Space after theorem head, ' ', or \newline
  {}%                                     % Theorem head spec (can be left empty, meaning `normal')
\theoremstyle{mystyle}
\newtheorem{definition}{Definition}

\newtheorem{assumption}{Assumption}
\usepackage{todonotes} % for missingfigure

\title{\LARGE\bf
Learning Object Relations with Graph Neural Networks

for Target-Driven Grasping in Dense Clutter
}
\author{Xibai Lou$^{1}$, Yang Yang$^{2}$, and Changhyun Choi$^{1}$% <-this % stops a space
\thanks{*This work was in part supported by the MnDRIVE Initiative on Robotics, Sensors, and Advanced Manufacturing.}% <-this % stops a space
\thanks{$^{1}$X. Lou and C. Choi are with the Department of Electrical and Computer Engineering, Univ. of Minnesota, Minneapolis, USA
        {\tt\small \{lou00015, cchoi\}@umn.edu}}%
\thanks{$^{2}$Y. Yang is with the Department of Computer Science and Engineering, Univ. of Minnesota, Minneapolis, USA {\tt\small yang5276@umn.edu}}%
}

\begin{document}

\maketitle
\thispagestyle{empty}
\pagestyle{empty}

%%%%%%%%%%%%%%%%%%%%%%%%%%%%%%%%%%%%%%%%%%%%%%%%%%%%%%%%%%%%%%%%%%%%%%%%%%%%%%%%
\begin{abstract}
Robots in the real world frequently come across identical objects in dense clutter. When evaluating grasp poses in these scenarios, a target-driven grasping system requires knowledge of spatial relations between scene objects (e.g., proximity, adjacency, and occlusions). To efficiently complete this task, we propose a target-driven grasping system that simultaneously considers object relations and predicts 6-DoF grasp poses. A densely cluttered scene is first formulated as a \textit{grasp graph} with nodes representing object geometries in the grasp coordinate frame and edges indicating spatial relations between the objects. We design a Grasp Graph Neural Network (G2N2) that evaluates the \textit{grasp graph} and finds the most feasible 6-DoF grasp pose for a target object. Additionally, we develop a shape completion-assisted grasp pose sampling method that improves sample quality and consequently grasping efficiency. We compare our method against several baselines in both simulated and real settings. In real-world experiments with novel objects, our approach achieves a 77.78\% grasping accuracy in densely cluttered scenarios, surpassing the best-performing baseline by more than 15\%. Supplementary material is available at \href{https://sites.google.com/umn.edu/graph-grasping}{https://sites.google.com/umn.edu/graph-grasping}.
\end{abstract}
\smallbreak

%%%%%%%%%%%%%%%%%%%%%%%%%%%%%%%%%%%%%%%%%%%%%%%%%%%%%%%%%%%%%%%%%%%%%%%%%%%%%%%%
\section{INTRODUCTION}

In recent years, robots have become increasingly prevalent in everyday life where environments are unstructured and partially observable. When retrieving a target object from such real-world scenarios, we frequently encounter a plethora of identical objects (e.g., grocery carts, tool boxes, warehouse storage, etc.). Humans understand spatial relations between scene objects and plan grasp poses accordingly for the task~\cite{feix2014analysis}. Moreover, our fingers occasionally knock away surrounding objects\textemdash this collision even facilitates grasping. When a robot faces the same situation, as illustrated in Fig. 1, we argue that an efficient grasping system should reason about spatial relations between objects to grasp a more accessible target while utilizing minor collisions.

Recent methods commonly solve this problem by cascading an object-centric grasp pose predictor with an analytic~\cite{mousavian2019graspnet} or data-driven collision checking module~\cite{murali2020clutteredgrasping}. While these two-stage approaches transform the problem into the well-studied single object grasping, they frequently struggle in the collision checking stage with imperfect and partial camera observations (i.e., fail to predict collisions with the occluded parts). Furthermore, because the two stages are carried out independently, these approaches overlook the fact that spatial relations between scene objects intrinsically affect the grasping success probability. In the aforementioned scenario, the absence of the knowledge of spatial relations between objects prevents the robot from efficiently grasping a target object.

\begin{figure}[t]
    \includegraphics[width=\linewidth]{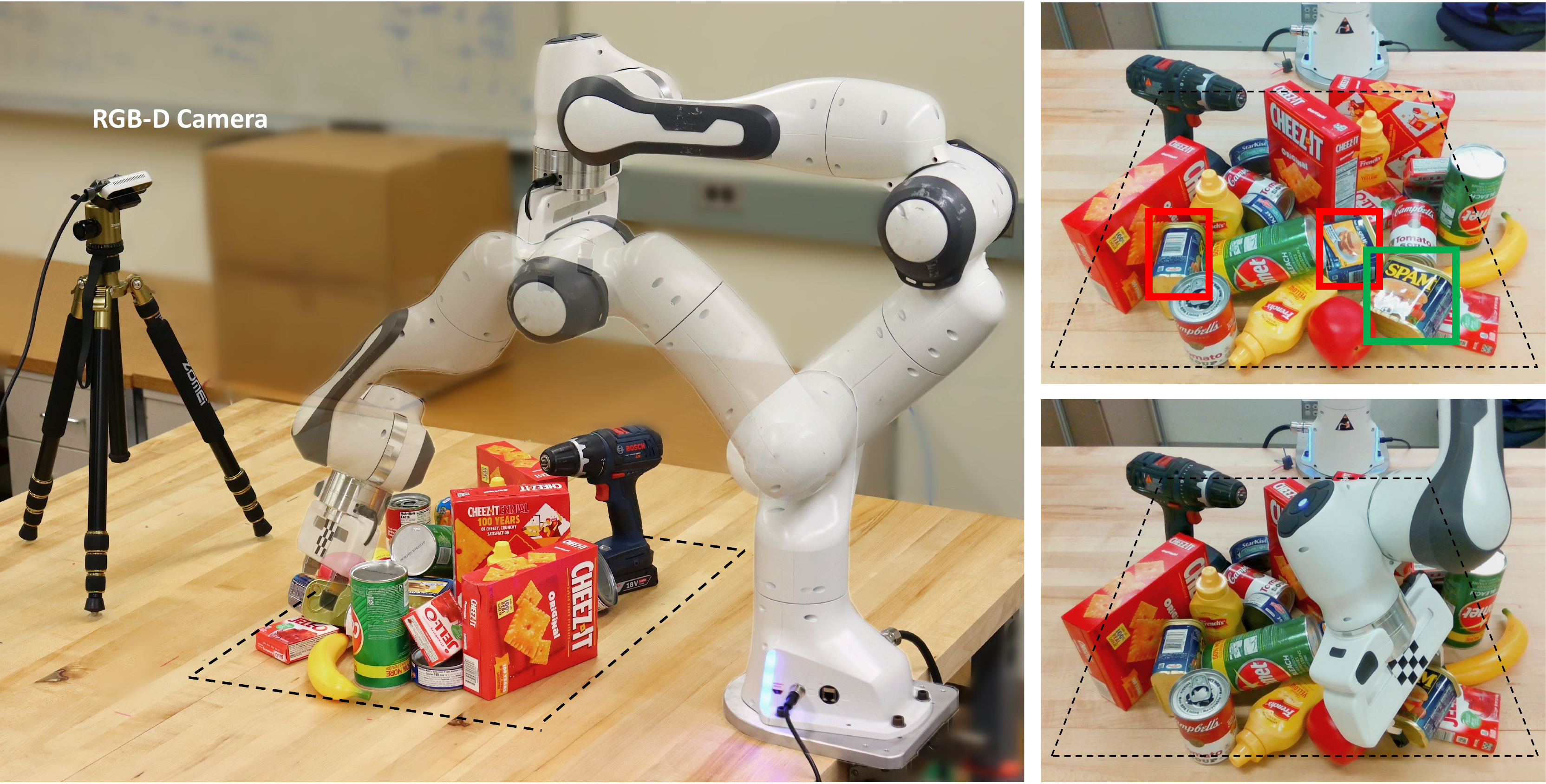}
    \caption{\textbf{Grasping one of the identical targets in dense clutter.}  Among the three meat cans, the one in the green bounding box is less surrounded by other objects. Grasping the more accessible target with a flexible pose is more likely to succeed. Reasoning about object relationships, the proposed target-driven grasping system predicts more feasible grasp poses in dense clutter.}
  \label{fig:cover}
  \vspace{-8pt}
\end{figure}

To fill the gap between the object-centric reasoning and the collision checking module, an end-to-end approach is desired for natural and coherent learning of the spatial relations between scene objects. Due to the vast 6-DoF action space and infinite object shapes and arrangements, learning directly from unstructured scene observations with traditional deep learning architectures (e.g., FCNs~\cite{long2015fully}, CNNs) does not generalize well. Graphs, on the other hand, have superior expressive capabilities since they explicitly model entities (e.g., objects, books, individuals, etc.) as nodes (containing descriptive features) and their relations as edges. Operating on graphs, Graph Neural Networks (GNNs)~\cite{4700287} are capable of efficiently capturing the hidden correlations between nodes~\cite{ZHOU202057}. In light of the recent advances in utilizing GNNs in robotics~\cite{garcia2019tactilegcn, wilson20a, murali2020taskgrasp}, we propose a 6-DoF target-driven grasping system with a Grasp Graph Neural Network (G2N2). A 3D Autoencoder first learns scene representations and encodes a 3D observation in a grasp coordinate frame to a graph with nodes representing object geometries and edges indicating their spatial relationships. Given a set of sampled poses, the G2N2 directly predicts the grasping success probabilities from the generated graphs, accounting for both grasp stability and object relations. Since the quality of sampled grasp candidates is critical to the efficiency of our grasping system, we additionally propose a shape completion-assisted sampling method that significantly improves sampling quality for partial geometries.

Despite sim-to-real gaps, our grasping system, which is entirely trained on synthetic datasets, generalizes well to the real world. We performed experiments in both simulated and real environments, in which our approach achieved a 75.25\% and 77.78\% grasping accuracy respectively for densely cluttered novel targets, outperforming the baselines by large margins. Our work is one of the early attempts that utilizes GNNs for a challenging 6-DoF target-driven robotic grasping task. The core technical contributions are:

\begin{itemize}
\item 
A novel Grasp Graph Neural Network (G2N2) that builds upon Graph Convolutional Networks~\cite{kipf2016gcn}. It learns to understand the spatial relations between scene objects and predicts feasible 6-DoF grasp poses for challenging robotic tasks such as retrieving a target object from several identical ones in dense clutter.
\end{itemize}

\begin{itemize}
\item 
A 6-DoF target-driven grasping system that contains a shape completion-assisted grasp pose sampling algorithm. The method improves sample efficiency for partially observable objects and leads to more effective data collection and training.
\end{itemize}

% \begin{itemize}
% \item 
% Extensive experiments on four comparable approaches in both simulated and real-world environments, where
% our work achieves superior results in all tasks. In particular, we surpass the current state-of-the-art performance by xx\%.
% \end{itemize}

\begin{figure*}[t]
    \centering
    \includegraphics[width=1.0\textwidth]{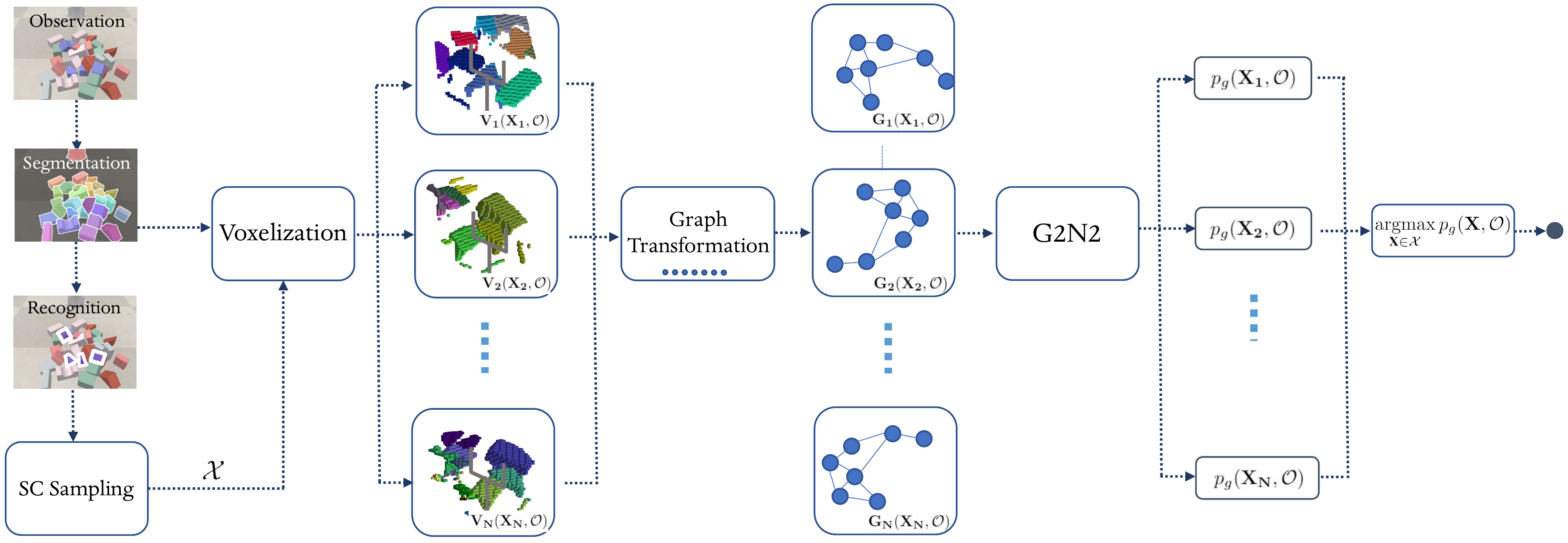}
    \caption{\textbf{Grasping pipeline.} Our approach first segments the RGB-D image and localizes the target objects, and then samples diverse 6-DoF grasp candidates for each target object. The object point clouds are transformed into each candidate's grasp coordinate frame and voxelized. Next, we apply graph transformation to construct grasp graphs, where the voxel grids are encoded into latent features and used as node embeddings. For a set of $N$ grasp candidates, the G2N2 evaluates the grasp graph for each candidate, and predicts an overall grasping score that simultaneously considers grasping stability and spatial relations between objects.}
    \label{fig:flowchart}
\vspace{-8pt}
\end{figure*}

\section{RELATED WORK}
Robotic grasping is a fundamental skill that enables further manipulation tasks~\cite{sahbani2012overview}, \cite{bohg2013data}. Recently, data-driven approaches~\cite{lenz2015deep, pinto2016supersizing} have applied Convolutional Neural Networks (CNNs) to learn grasping in 3-DoF action space~\cite{zeng2018learning, liang2019knowledge, yang2021attribute}. Meanwhile, other works learn the more flexible 6-DoF grasping~\cite{gualtieri2016high, ten2017grasp, liang2019pointnetgpd}. Choi et al. employed a 3D CNN to classify 24 pre-defined orientations from object voxel occupancy grids~\cite{choi2018learning} and Lou et al. extended the work to full 6-DoF and introduced reachability awareness~\cite{lou2020learning}. Based-on the PointNet++~\cite{qi2017pointnet} architecture, Mousavian et al. learned a variational sampling method and grasp pose evaluator to detect stable 6-DoF poses~\cite{mousavian2019graspnet}. When grasping a target from clutter, these approaches often rely on an analytical~\cite{mousavian2019graspnet, liang2019knowledge} or data-driven collision checking module~\cite{murali2020clutteredgrasping, lou2021collision}. Consequently, these approaches overlook the importance of spatial relations between objects to grasping results. Unlike the previous target-driven methods, we focus on a more demanding task, retrieving a pre-defined target from several identical objects in dense clutter, where knowledge of object relationships is essential. We use graphs to learn deep relations between numerous objects because they clearly define the entities and their relations in a compact and effective form~\cite{xu2018powerful}.

Recently, Graph Neural Networks (GNNs)~\cite{4700287} have been applied to multiple disciplines that necessitates processing rich relation information, including social networks~\cite{wu2018socialgcn}, citation systems~\cite{velivckovic2017graph}, natural language processing~\cite{yao2019graph}, and particle dynamics~\cite{sanchez2020learning}. Following the success of GNNs in the these fields, robotics community has shown growing interests. By structuring a neural network to pass information between adjacent nodes, Murali et al. use Graph Convolutional Networks (GCNs)~\cite{kipf2016gcn} to reason about the relationships between the object classes and the tasks~\cite{murali2020taskgrasp}. Wilson et al. learns representations that enable GNN to infer how to manipulate a group of objects~\cite{wilson20a}. Garcia-Garcia et al. proposed TactileGCN~\cite{garcia2019tactilegcn}, which predicts grasping stability from a graph of tactile sensor readings. In contrast, we propose a target-driven approach that employs a Grasp Graph Neural Network (G2N2) to capture these relations.

Aside from data-driven models, sampling methods are equally critical to grasp pose detection methods~\cite{ten2017grasp, liang2019pointnetgpd, lou2020learning}. Given a single-view observation, reconstructing the missing geometry will naturally enhance sampling efficiency. 3D shape completion from partial observations has been studied in computer vision and graphics~\cite{shapenet2015, FirmanCVPR2016, 7298801} and applied to robotics. Varley et al. trained a shape completion module with 3D CNNs \cite{varley2017shape} and performed analytical grasp planning on the output shape. Rather than solely relying on the shape completion module, we use it only for grasp pose sampling. Hence, errors in shape completion that lead to unsuccessful grasps will be learned by the G2N2 and penalized during evaluation.

\section{PROBLEM FORMULATION}
\label{section:problem}
The goal of our work is to investigate target-driven robotic grasping in dense clutter, where several identical targets may be present. Besides grasp pose stability, this task also demands intensive reasoning about spatial relations between scene objects (e.g., relative locations, occlusions, etc.) when predicting the most feasible grasp pose. We define our problem as follows:

\begin{definition}
Given a query image, the robot needs to retrieve the same item from the dense clutter where several identical target objects may be present.
\end{definition}

With a single-view camera observation $\mathcal{O}$ of the open world, the challenge arises from:

\begin{assumption}
The objects are possibly unknown (i.e., novel objects) and partially observable (e.g., object occlusions and imperfect sensors).
\end{assumption}

The partial observations inevitably increase collisions during grasping. A robot that understands spatial relations between objects can adapt to collisions. To learn such relationships with GNNs, we formulate a graph representation: 

\begin{definition}
A \textbf{grasp graph} includes grasp pose information, geometric features of objects, and the spatial relations between these objects.
\end{definition}

However, objects far away from the target are less relevant to our task, therefore, we assume that:

\begin{assumption}
The objects that are distant from a target object have limited influence on the grasping success rate.
\end{assumption}

Given a set of uniformly sampled grasp candidates $\mathcal{X} \subset SE(3)$, a grasp graph $\mathbf{G}(\mathbf{X}, \mathcal{O})$ is constructed from grasp candidate $\mathbf{X} \in \mathcal{X}$ and the scene observation $\mathcal{O}$. The grasp pose $\mathbf{X}$ is subject to a binary-valued metric $\mathcal{S}_g(\mathbf{X})\in$ \{0,1\} where $\mathcal{S}_g = 1$ indicates that the grasp is successful. The grasping success probability $p_g(\mathbf{X}, \mathcal{O}) = Pr(\mathcal{S}_g = 1|\mathbf{G(\mathbf{X}, \mathcal{O})})$.

\section{PROPOSED APPROACH}
\label{section:proposed_approach}

\begin{algorithm}[b]
\caption{GNN Target-driven Grasping}
\label{algo:grasping}
\hspace*{\algorithmicindent} \textbf{Input: }RGB-D Image $\mathcal{I}$, GNN $\mathcal{N}_g$, Shape Encoder $\phi_v$, Edge Encoder $\phi_e$\\
\hspace*{\algorithmicindent} \textbf{Output: } grasp pose $\mathbf{X}_g\in{SE(3)}$ 
\begin{algorithmic}[1]
\State $\mathcal{O} \gets \texttt{Segmentation($\mathcal{I}$)}$
\State $\mathcal{P}\gets \texttt{BackProjection($\mathcal{O}$)}$
\State $\mathcal{O}_t \gets \texttt{Recognition($\mathcal{O}$)}$
\For{${\mathbf{O}}\in\mathcal{O}_t$}
    \State $\mathbf{P}\gets \texttt{BackProjection($\mathbf{O}$)}$
    \State $\mathcal{X} \gets \texttt{GraspPoseSampling($\mathbf{P}$)}$
    \For{${\mathbf{X}}\in\mathcal{X}$}
        \State $\mathbf{v}_i \gets \texttt{VoxelTransformation}(\mathcal{P}, \mathbf{X})$
        \State $\mathbf{x}_i, \mathbf{y}_{ij} \gets \phi_v(\mathbf{v}_i), \phi_e(\mathbf{e}_{ij})$
        \State $\mathcal{G} \gets \texttt{GraphConstruction($\mathbf{x}_{i}, \mathbf{y}_{ij}$)}$
        \State $p_g \gets \texttt{$\mathcal{N}_g$.Feedforward}(\mathcal{G})$
    \EndFor
\EndFor
\State $\mathbf{X}_g \gets \argmax_{\mathbf{X} \in \mathcal{X}} p_g(\mathbf{X}, \mathcal{O})$
\State $\texttt{Grasp}(\mathbf{X}_g)$
\end{algorithmic}
\end{algorithm}

The proposed 6-DoF grasping system efficiently retrieves one of the identical targets from dense clutter. This task is challenging as it demands intensive reasoning about the spatial relations between scene objects. To incorporate grasp pose information into the latent features of the scene, the proposed grasping system first transforms the object point clouds into the grasp coordinate frame, and extracts the geometric features from each object with a 3D Convolutional Autoencoder. Using the features as node embeddings, we first construct a \textit{grasp graph} representing the grasp pose, object geometries, and spatial relations between them. Next, we train an end-to-end model, Grasp Graph Neural Network (G2N2), on a synthetic dataset of grasp graph to predict the grasping success probability. During testing, we employed a shape completion-assisted sampling method to propose diverse grasp candidates, then the G2N2 evaluates all grasp graphs and select the highest-scored grasp pose to execute. The system is depicted in Fig.~\ref{fig:flowchart} and delineated in Algorithm~\ref{algo:grasping}.

\subsection{Pose to Graph Transformation} 
\label{subsec:grasp_graph}
Learning an effective graph representation is critical to the proposed G2N2. Our grasping system formulates the observation and a grasp candidate into a \textit{grasp graph} representation containing spatial features, defined in Sec.~\ref{section:problem}. For a scene of $N$ objects, a grasp graph $\mathcal{G}$ contains a tuple of two sets $\mathcal{G} = (\mathcal{V}, \mathcal{E})$, where $\mathcal{V} = {\mathbf{v}_1, \mathbf{v}_2, ..., \mathbf{v}_N}$ and $\mathcal{E} \in \mathcal{V} \times \mathcal{V}$ are the sets of nodes and edges, respectively. The graph is directed and $\mathbf{e}_{ij}$ is an edge from $\mathbf{v}_i$ to $\mathbf{v}_j$. A node $\mathbf{v}_i$ is associated with node features $\mathbf{x}_i \in \mathbb{R}^D$ and an edge $\mathbf{e}_{ij}$ with edge features $\mathbf{y}_{ij} \in \mathbb{R}^C$, where $D = C = 128$ represents the initial dimensions of node and edge features.

We adopt the UOIS~\cite{xie2019uois} for segmentation and the Siamese network-based target matching in~\cite{lou2020learning} for localizing targets in the scene observation $\mathcal{O}$. Then object point clouds are transformed by the 6-DoF grasp pose $\mathbf{X}$ into the grasp coordinate frame (i.e., the axes are aligned with the gripper reference frame and the tool center point becomes the origin). This transformation encodes the grasp pose within the point clouds. Since distant objects have little impact on the prediction, we set an effective workspace of ${(0.2m)}^3$ around the grasping point. The valid object point clouds are voxelized to a set of voxel grids $\mathbf{v}_i$ that represents the nodes in the grasp graph. Next, the 128-dimensional node features $\mathbf{x}_i$ are extracted using a 3D Convolutional Autoencoder $\phi_v$: Conv3D(1, 32, 5) $\rightarrow$ ELU $\rightarrow$ Maxpool(2) $\rightarrow$ Conv3D(32, 32, 3) $\rightarrow$ ELU $\rightarrow$ Maxpool(2) $\rightarrow$ FC($32 \times 6 \times 6 \times 6$, 128). The grasp graph is fully connected, and $\mathbf{e}_{ij} = (d_{ij}, z_{ij})$ where $d_{ij} \in \mathbb{R}^3$ is a 3D vector pointing from source node $\mathbf{v}_i$ to target node $\mathbf{v}_j$ and $z_{ij} \in \mathbb{R}^{+}$ is their Euclidean distance. The grasp graph $\mathcal{G}$ is constructed such that each node is composed of the latent geometric feature $\mathbf{x}_i = \phi_v(\mathbf{v}_i)$ and each edge is composed of spatial feature $\mathbf{y}_{ij} = \phi_e(\mathbf{e}_{ij})$ where $\phi_e$ is a Multi-Layer-Perceptron (MLP) that extracts edge features from edge attributes. 

% Graphs define abstract concepts like object geometries and spatial information as nodes and edges. With such strong expressiveness, 

\subsection{Grasp Graph Neural Network}
GNNs specialize in discovering the underlying relationships between nodes. In graph learning, a non-linear function $\mathcal{F}$ takes a graph representation $\mathcal{G}$ as input and maps it to $\mathcal{G}'$ with updated node features~\cite{li2020deepergcn}. Graph updates in GNNs combine an arbitrary number of these nonlinear functions, and produce node features containing learned knowledge about the neighborhoods. We define the GNN structure by finding a set of operators parameterized by the grasp graph dataset such that the output graph representation is useful for our grasp pose prediction task. The graphs are updated as follows:

\vspace{-8pt}
\begin{align}
\label{eq:F}
\mathcal{G}^{(l+1)} = \mathcal{F}^{(l)} \left( \mathrm{ReLU} \left( \mathcal{G}^{(l)}(\mathcal{V}^{(l)}, \mathcal{E}^{(l)}) \right) \right), \forall l \in [0, L]
\end{align}

Given a grasp graph, we aggregate the node features $\mathbf{x}_{i}$ following the intuitions in ~\cite{li2020deepergcn} by incorporating a generalized \textit{SoftMax()} aggregator $\sigma(\cdot)$. Experimentally shown in~\cite{li2020deepergcn}, the learned parameters in $\sigma(\cdot)$ define a more effective aggregation scheme than \textit{mean()} or \textit{max()} alone for the learning task. Eq.~\ref{eq:GEN} describes the message-passing function for updating latent features $\mathbf{x}_i$ of node $\mathbf{v}_i$ with its neighbors:
\begin{figure}[t]
    \centering
    \includegraphics[width=1.0\linewidth]{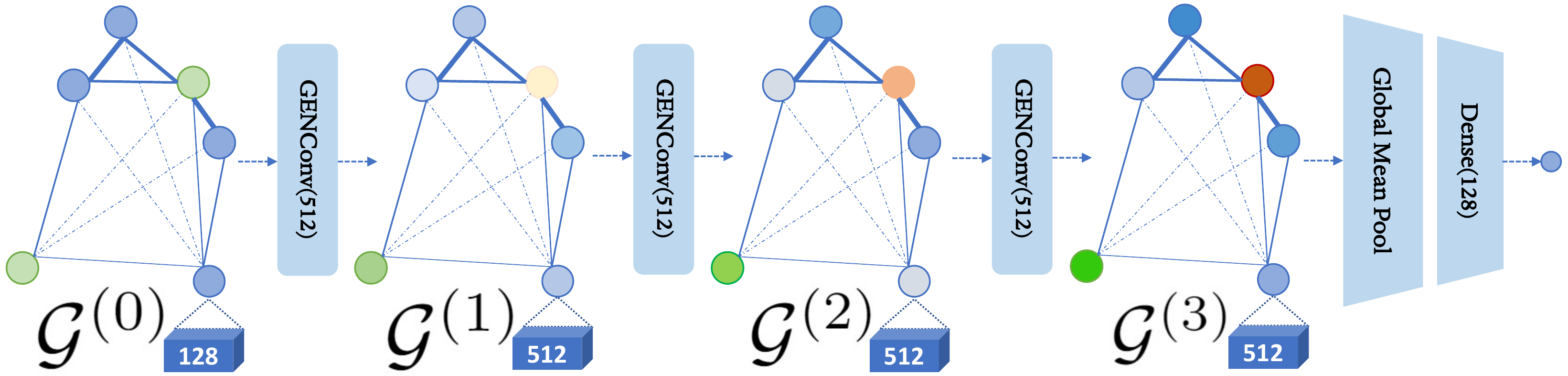}
    \caption{\textbf{Grasp Graph Neural Network structure.} The input graph has 128-dimensional node features and edge features. The GENConv layers transform the node features to 512 units. Graph-level features are extracted with a global-mean pooling layer, and then feed to a post message-passing MLP to infer the grasping success probability $p_g(\mathbf{X}, \mathcal{O})$.}
    \label{fig:gnn_structure}  
    \vspace{-21pt}
\end{figure}
\begin{align}
\label{eq:GEN}
\mathbf{x}_i^{\prime} = \mathrm{\phi} \left( \mathbf{x}_i + \mathrm{\sigma} \left( \left\{\mathrm{ReLU} \left( \mathbf{x}_j + \mathbf{y}_{ji} \right) +\epsilon: j \in \mathcal{N}(i) \right\} \right)\right)
\end{align}

Our network sequentially connects three GNN layers ($L = 2$ in Eq.~\ref{eq:F}). Each layer has node feature $\mathbf{x}_i \in \mathbb{R}^H$ where $H = 512$. Intuitively it convolves the 3rd-order neighbors of every node, capturing copious spatial relations. After the graph operations, the graph-level output features are extracted from the final graph $\mathcal{G}^{L+1}$ with a global-mean pooling function $f(\cdot)$, then we compute the grasping success probability $p_g(\mathbf{X}, \mathcal{O}) = \mathrm{\phi_o}(f(\mathcal{G}^{L+1}))$ where $\phi_o$ represents a post message-passing MLP. Detailed network structure is shown in Fig.~\ref{fig:gnn_structure}. At a high-level, each grasp graph represents a candidate, and the G2N2 consumes the grasp graphs and estimates the grasping success probability $p_g(\mathbf{X}, \mathcal{O}) = Pr(\mathcal{S}_g = 1|\mathbf{G(\mathbf{X}, \mathcal{O})})$ in an end-to-end fashion. Finally, we select the highest scored grasp pose.  
% \[z = \argmax_{\textbf{X} \in \mathcal{X}} p_g(\mathbf{X}, \mathcal{O})\]

\subsection{Grasp Sampling}
\begin{figure}[t]
    \includegraphics[width=\linewidth]{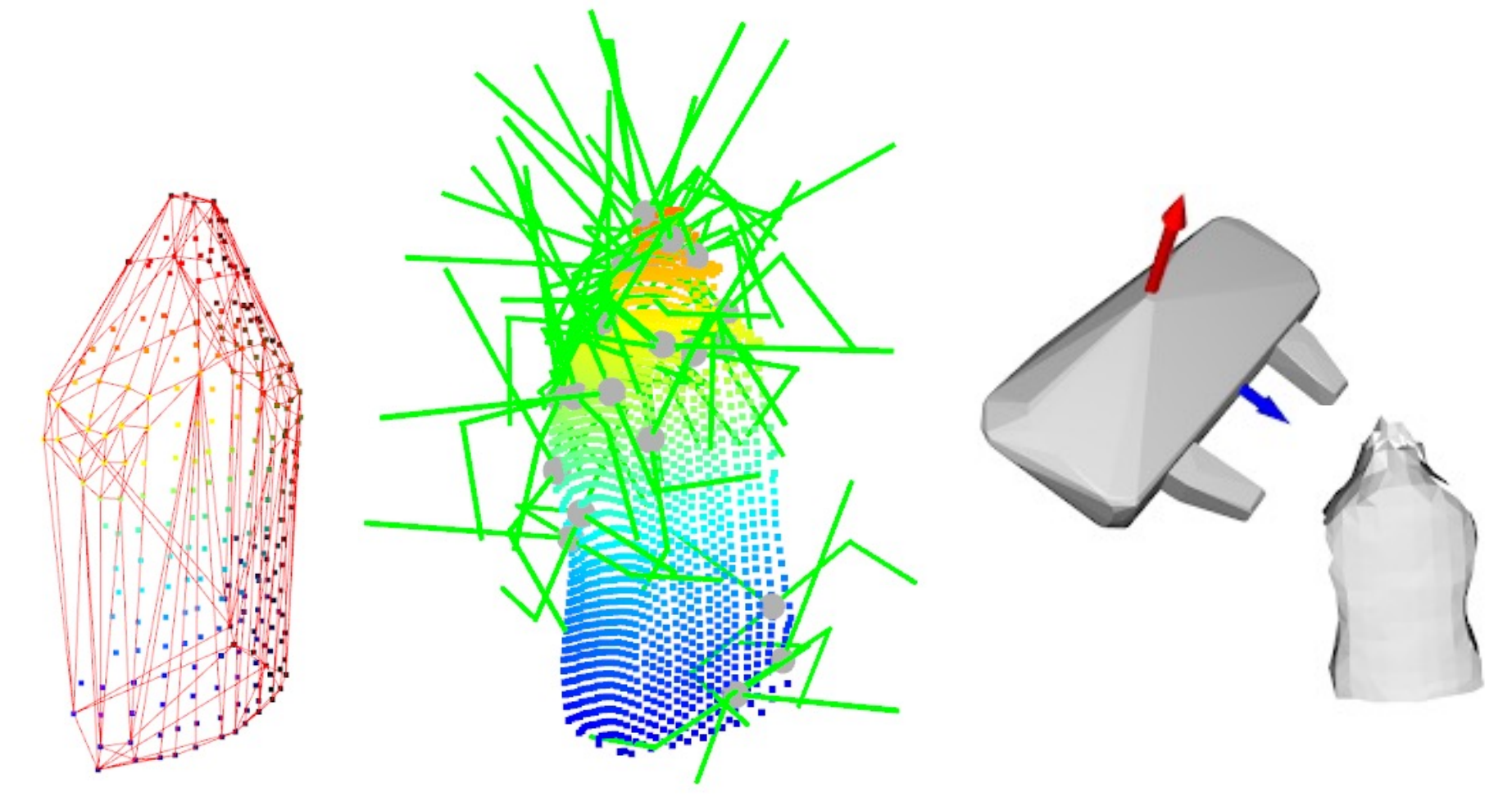}
    \caption{\textbf{Shape completion-assisted 6-DoF grasp sampling.} With additional constrains and the shape completion module, the algorithm keeps the sampling diversity, while achieving better efficiency.}
 \label{fig:sampling}
 \vspace{-8pt}
\end{figure}

Geometric-based sampling generates grasp poses constrained by the surface normal and local direction of the principal curvature~\cite{ten2017grasp, liang2019pointnetgpd}. Uniform random sampling, on the other hand, covers a larger action space~\cite{lou2020learning}, however, the sample quality often suffers under partial observation because grasp candidates may collide with the occluded geometries. To reduce these low quality samples, we first infer the occluded geometry of the object with a 3D CNN shape completion module: Conv3D(1, 32, 5) $\rightarrow$ ELU $\rightarrow$ Maxpool(2) $\rightarrow$ Conv3D(32, 32, 3) $\rightarrow$ ELU $\rightarrow$ Maxpool(2) $\rightarrow$ FC($32\times6\times6\times6$, 128) $\rightarrow$ ReLU $\rightarrow$ FC(128, $32\times6\times6\times6$) $\rightarrow$ Maxunpool(2) $\rightarrow$ ConvTranspose3D(32, 32, 3) $\rightarrow$ Maxunpool(2) $\rightarrow$ ConvTranspose3D(32, 1, 5). The sampled grasp poses are constrained such that the angle between their approach directions and the object surface normal are less than $\pi/4$ (i.e., the gripper will not slide over the object). The process is depicted in Fig.~\ref{fig:sampling}. The key differences between our approach and other comparable works~\cite{varley2017shape, ten2017grasp, liang2019pointnetgpd} are 1) the diversity of the sampled grasp candidates is not compromised while more plausible grasp samples contribute to a more balanced dataset, 2) grasp candidates of higher quality also improve grasping success rate, especially under heavy occlusion (i.e., single viewed observation, object-object occlusions), and 3) our system uses shape completion only in the sampling stage, therefore, errors have negligible impact on grasping prediction.

\subsection{Data Collection and Training}
We first collect 3,000 depth images of the randomly positioned training objects (toy blocks of various geometric shapes) and voxelize the scene point clouds into occupancy grids. The dataset is prepared for representation learning with the 3D Convolutional Autoencoder described in Sec.~\ref{subsec:grasp_graph}. Using the same dataset, a shape completion module for sampling is trained by regressing the input partial voxel grid against the complete ground truth voxel grid, obtained by transforming the full mesh from the canonical pose.

The grasping simulation environment is built in CoppeliaSim v4.0 with the Bullet physics engine version 2.83. We randomly place training objects on the tabletop to form a dense clutter, and iteratively attempt 32 grasps for each arrangement. The grasp graph dataset contains 3,000 scenes and 96,000 grasps in total. After transforming the scene observations to a grasp graph dataset into a lower dimension, we implemented and trained the G2N2 in PyTorch Geometric with the binary cross-entropy loss and Adam optimizer until validation accuracy plateaued.

\section{EXPERIMENTS}

We experiment in diverse settings in both simulated and real-world environments. The experiments are designed to address the following questions: i) How effective is the proposed approach in retrieving one of the identical targets in dense clutter?, ii) How the proposed approach performs compared to other two-stage methods?, and iii) Does the proposed approach generalize to more challenging and novel scenarios?

To answer those questions, we test the proposed approach and three comparable works: 
\begin{enumerate}
    \item \textbf{\texttt{GSP}} stands for Grasp Stability Predictor~\cite{lou2020learning}\footnote{Note that the acronym GSP refers to the 3D CNN grasping module in~\cite{lou2020learning}, defined here for a concise reference.}, an object-centric 6-DoF grasp pose detection method that uniformly samples grasp candidates on the object point clouds and predicts their grasping success probabilities with 3D CNNs.
    \item \textbf{\texttt{6GN}} is 6-DoF GraspNet \cite{mousavian2019graspnet} that learns to sample diverse 6-DoF poses from the latent variables of a Variational Autoencoder. The candidates are evaluated by a PointNet++~\cite{qi2017pointnet} based network and further enhanced by gradient-based refinement.
    \item \textbf{\texttt{GSP+AS}} enhances GSP with an analytical search algorithm. It measures the number of voxels around each target object and assigns clutteredness scores that are proportional to voxel numbers. 
\end{enumerate}

For a fair comparison, the same target object masks are provided to each method and 100 grasp candidates per object are generated following their sampling methods. Our approach considers scene objects when evaluating grasps candidates, therefore it does not require additional collision checking. For the other object-centric methods, Flexible Collision Library (FCL)~\cite{6225337} is used to determine if the robot mesh is in collision with the scene mesh, which is reconstructed using the Marching Cubes algorithm~\cite{marchingcubes}. The highest scored collision-free grasp pose will be executed. We define the grasping accuracy as $ \frac{\text{\# of successful grasps}}{\text{\# of proposed grasps}}$. A grasp is successful only if the robot gripper successfully reaches the object and lifts the object by 15 cm. If no grasp pose is found among the sampled candidates, the grasp is also considered unsuccessful.

\subsection{Simulation Experiments}

\begin{figure}[t]
  \begin{subfigure}{0.154\textwidth}
    \includegraphics[width=\textwidth]{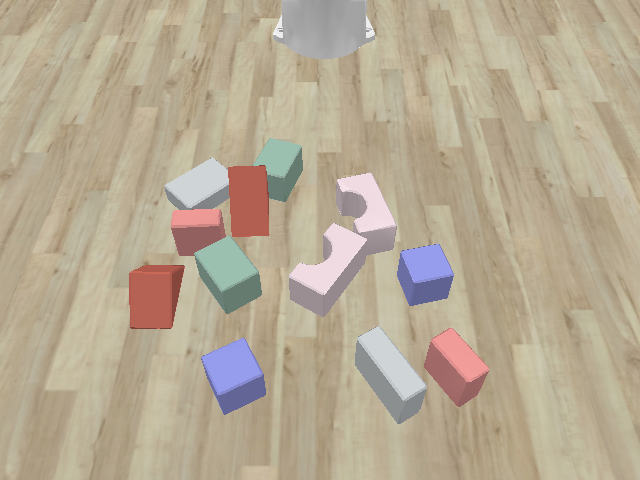}
    \caption{Block-Normal}
  \end{subfigure}
  \begin{subfigure}{0.154\textwidth}
    \includegraphics[width=\textwidth]{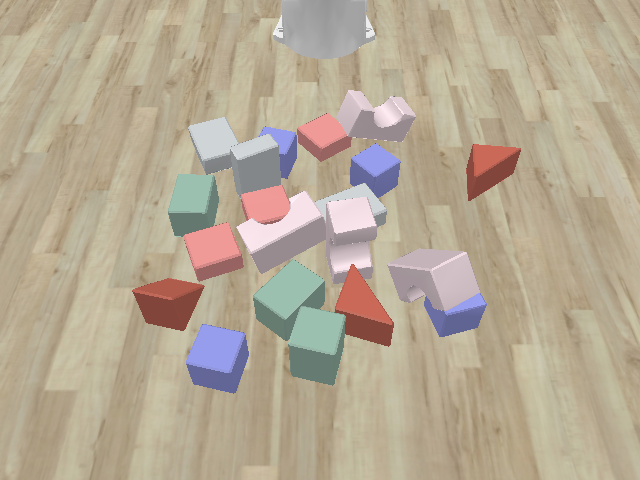}
    \caption{Block-Medium}
  \end{subfigure}
    \begin{subfigure}{0.154\textwidth}
    \includegraphics[width=\textwidth]{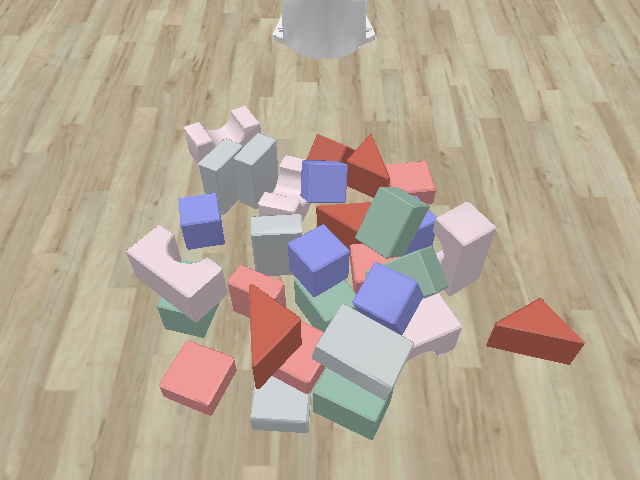}
    \caption{Block-Dense}
  \end{subfigure}
    \caption{\textbf{Block Clutter in simulation.} The tests are designed with an increasing difficulty from normal to densely cluttered.}
 \label{fig:sim_block}
 \vspace{-8pt}
\end{figure}

Our simulated environment includes a Franka Emika Panda robot arm and various test objects. A single-view RGB-D observation is taken with a simulated camera. The first set of experiments, shown in Fig.~\ref{fig:sim_block}, tests the different approaches with block clutter of increasing density. One set of toy blocks includes seven objects of different shapes. \textit{Block-Normal} is similar to the training settings where two sets of objects are randomly placed on the tabletop; \textit{Block-Medium} increases the clutter density by adding another set of objects; \textit{Block-Dense} adds 5 sets of objects (35 objects in total) onto the same workspace, creating a dense clutter. Each method runs 101 times and the results are summarized in Table~\ref{tab:sim_block}. 6GN occasionally fails for smaller objects, but all methods are effective in \textit{Block-Normal} since objects are lightly cluttered. As the density increases, GSP and 6GN perform poorly  as they often try to grasp less accessible ones. They are inevitably more prone to disastrous collisions due to occluded geometries. Although GSP+AS is able to select less cluttered objects analytically, it struggles in \textit{Block-Dense}, showing that the analytical clutteredness is largely limited under heavy occlusions. The proposed approach with G2N2 (Ours) performs best in all categories since it takes into account spatial relations between objects. We achieve a 84.16\% success rate in the \textit{Block-Dense} scenario, outperforming the next best baseline by 19.72\%.

% \begin{figure}[t]
% \label{fig:simres_block}
% \centering
% \includegraphics[width=0.49\textwidth]{images/grasp_acc_block.pdf}
% \includegraphics[width=0.49\textwidth]{images/grasp_acc_novel.pdf}
% \caption{\textbf{Experiment Results in Simulation.} G2N2 achieved a 89.10\% grasping accuracy, suggesting it is the most effective}.
% \end{figure}
\begin{table}[b]
\caption{Simulation Results in Block Clutter}
\label{tab:sim_block}
\begin{center}
\begin{tabular}{|c||c||c||c||c||c|}
\hline
  & GSP & 6GN & GSP+AS & Ours\\
\hline
Block-Normal & 91.09 & 87.13 & 92.08 & \textbf{93.06}\\
\hline
Block-Medium & 84.16 & 76.24 & 87.13 & \textbf{90.10}\\
\hline
Block-Dense & 61.39 & 57.43 & 70.30 & \textbf{84.16}\\
\hline
\end{tabular}
\end{center}
\end{table}

\begin{figure}[t]
\begin{subfigure}{0.154\textwidth}
\includegraphics[width=\textwidth]{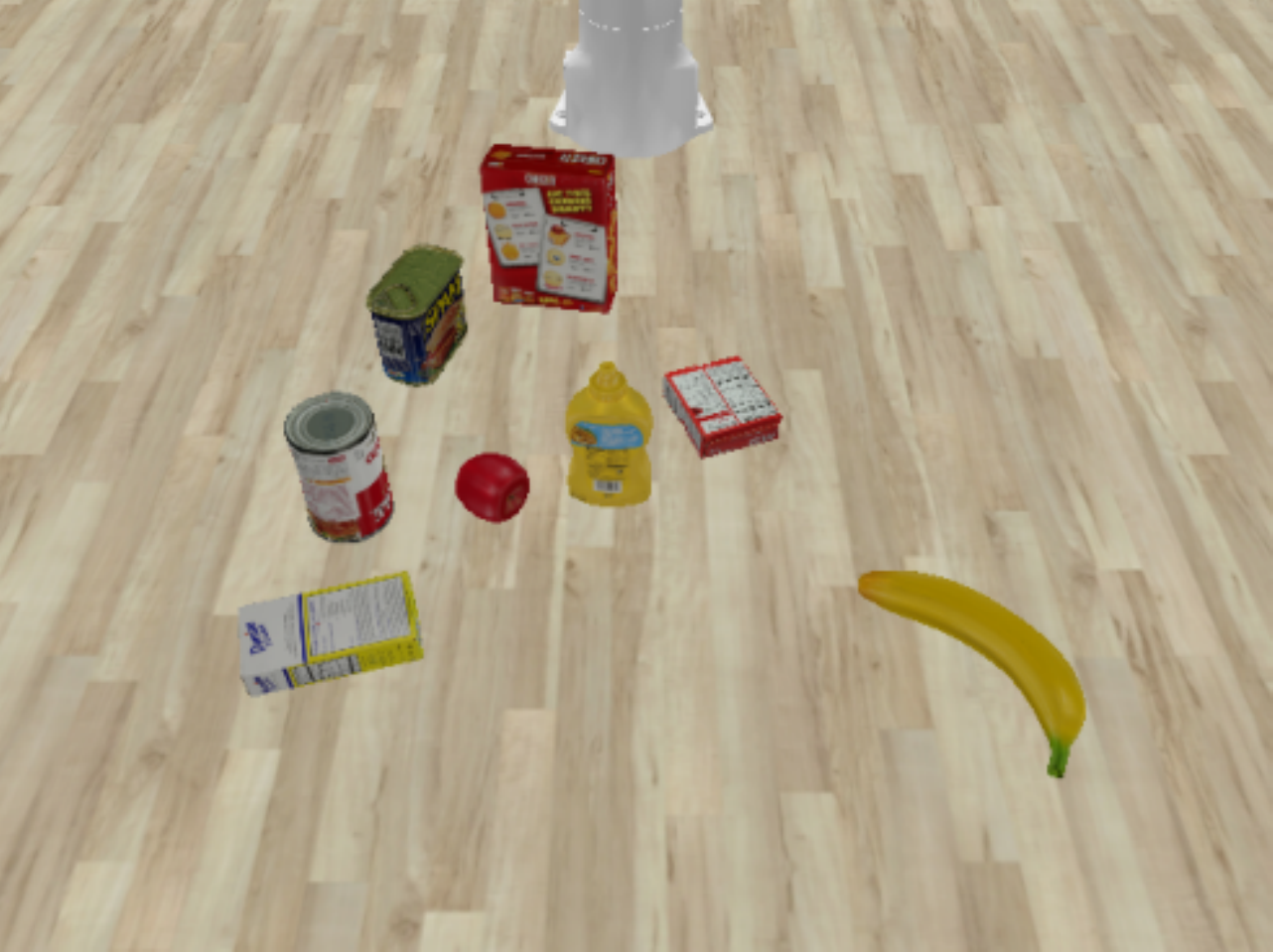}
\caption{Novel-Single}
\end{subfigure}
\begin{subfigure}{0.154\textwidth}
\includegraphics[width=\textwidth]{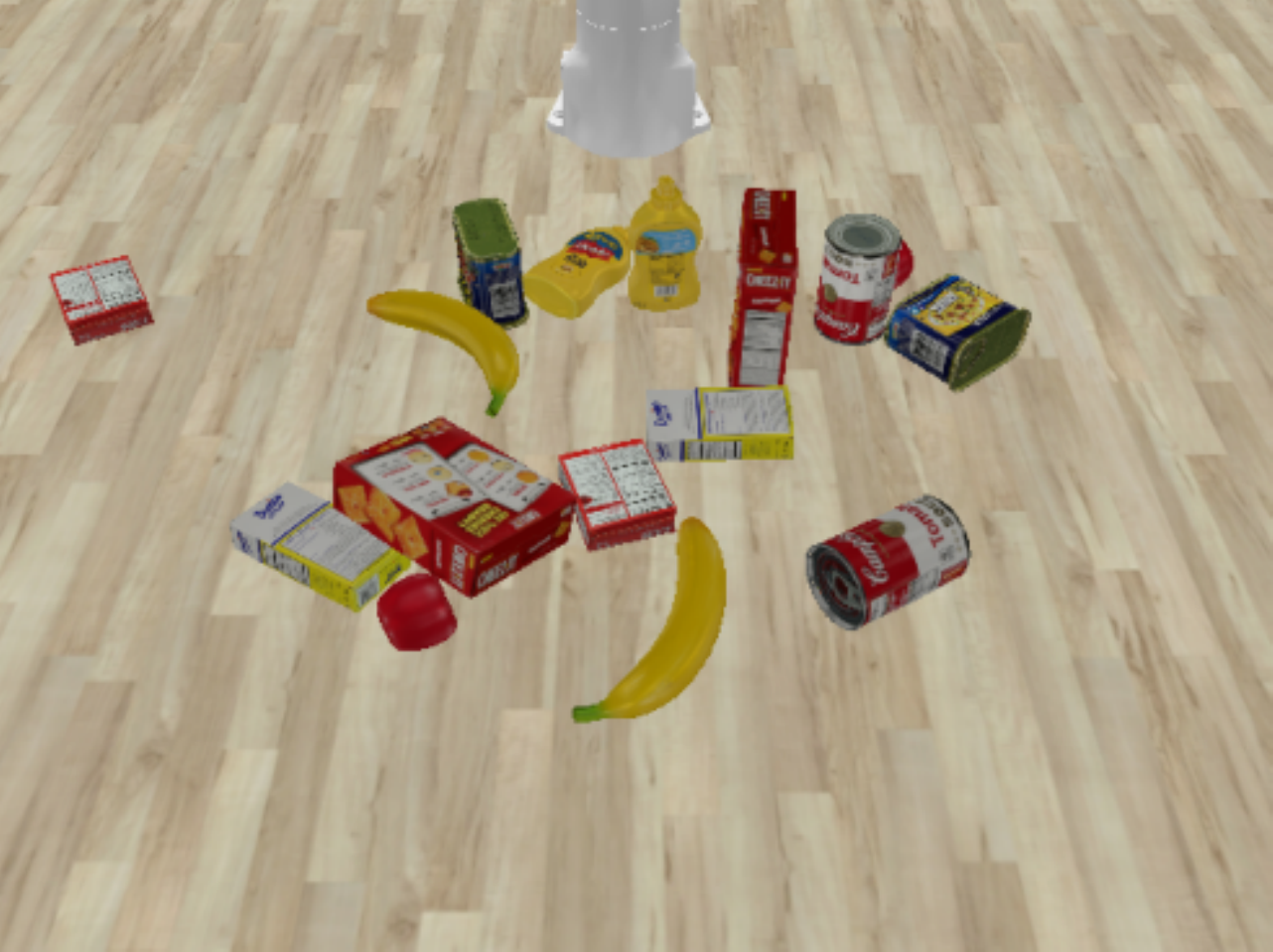}
\caption{Novel-Medium}
\end{subfigure}
\begin{subfigure}{0.154\textwidth}
\includegraphics[width=\textwidth]{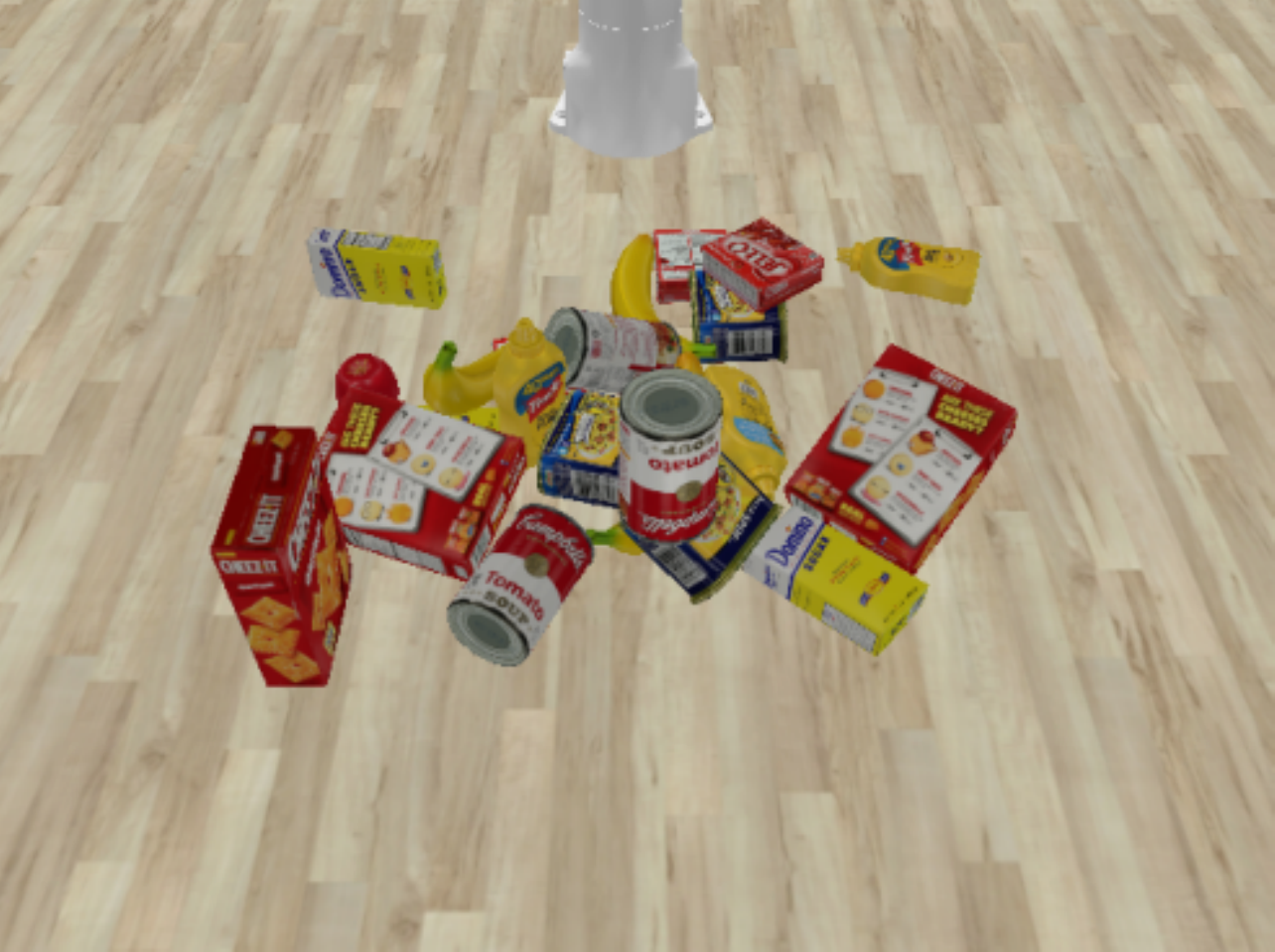}
\caption{Novel-Dense}
\end{subfigure}
\caption{\textbf{Novel scenes in simulation.} (a) consists of novel objects that are \textit{distinct} from our training set, (b) and (c) both contain identical target objects, but (c) introduces more occlusions.}
\label{fig:sim_novel}
\vspace{-8pt}
\end{figure}

\begin{table}[b]
\caption{Simulation Results in Novel Scenes}
\label{tab:sim_novel}
\begin{center}
\begin{tabular}{|c||c||c||c||c||c|}
\hline
  & GSP & 6GN & GSP+AS & Ours\\
\hline
Novel-Single & 76.24 & 83.16 & 76.24 & \textbf{87.13}\\
\hline
Novel-Medium & 66.34 & 74.25 & 71.29 & \textbf{84.16}\\
\hline
Novel-Dense & 54.45 & 63.37 & 70.30 & \textbf{81.19}\\
\hline
\end{tabular}
\end{center}
\end{table}

We further experiment in more demanding settings. \textit{Novel-Single} includes 8 objects from the YCB dataset~\cite{7251504} that are distinct from the training objects, focusing on testing the generalizability to novel objects. The friction coefficients heavily influence the outcome, and we use the common value according to each object's material. \textit{Novel-Medium} and \textit{Novel-Dense} double and triple the object numbers, respectively, increasing the density of the scenes. The scenes are illustrated in Fig.~\ref{fig:sim_novel}. Each method is tested 101 times, and the results are compiled in Table~\ref{tab:sim_novel}. Our approach achieves the highest 81.19\% grasping accuracy in \textit{Novel-Dense}, suggesting that it is generalizable to completely different settings that include novel objects.

The ablation studies are carried out in the \textit{Block-Dense} scenario to closely examine each component, summarized in Table~\ref{tab:sim_ab}. All other modules and parameters are controlled except the following alterations. \textit{noSC} uses the random sampling previously employed in~\cite{lou2020learning} instead of our new shape completion-assisted method, and it shows discouraging performance. \textit{noMsg} stops message passing in our network (equivalent to not performing any graph operations) and results in lower performance, suggesting that G2N2 clearly learns to aggregate node features $\mathbf{x}_i$ effectively. \textit{FCN} attempts to learn $p_g(\mathbf{X}, \mathcal{O})$ from the encoded features with fully connected layers but generalizes poorly. Compared to \textit{GCN} that uses Graph Convolutional Networks~\cite{kipf2016gcn} to learn graph representations, the G2N2 it is more effective. 

\begin{table}[b]
\caption{Ablation Study in Block-Dense}
\label{tab:sim_ab}
\begin{center}
\begin{tabular}{|c||c||c||c||c||c|}
\hline
  & noSC & noMsg & FCN & GCN & Ours\\
\hline
Grasp Acc. & 75.25 & 65.35 & 51.49 & 80.20 & 84.16\\
\hline
\end{tabular}
\end{center}
\end{table}

\subsection{Real-robot Experiments}

\begin{figure}[t] 
\begin{subfigure}{0.154\textwidth}
\includegraphics[width=\textwidth]{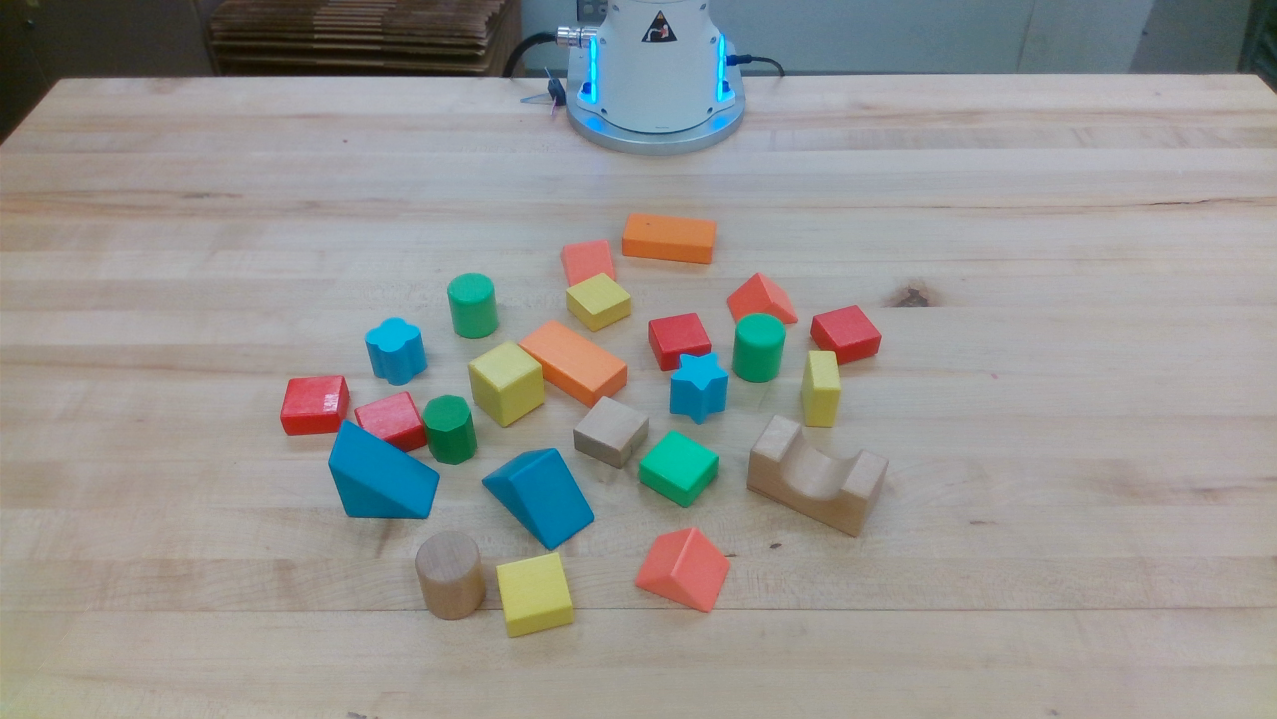}
\caption{Block-Normal}
\end{subfigure}
\begin{subfigure}{0.154\textwidth}
\includegraphics[width=\textwidth]{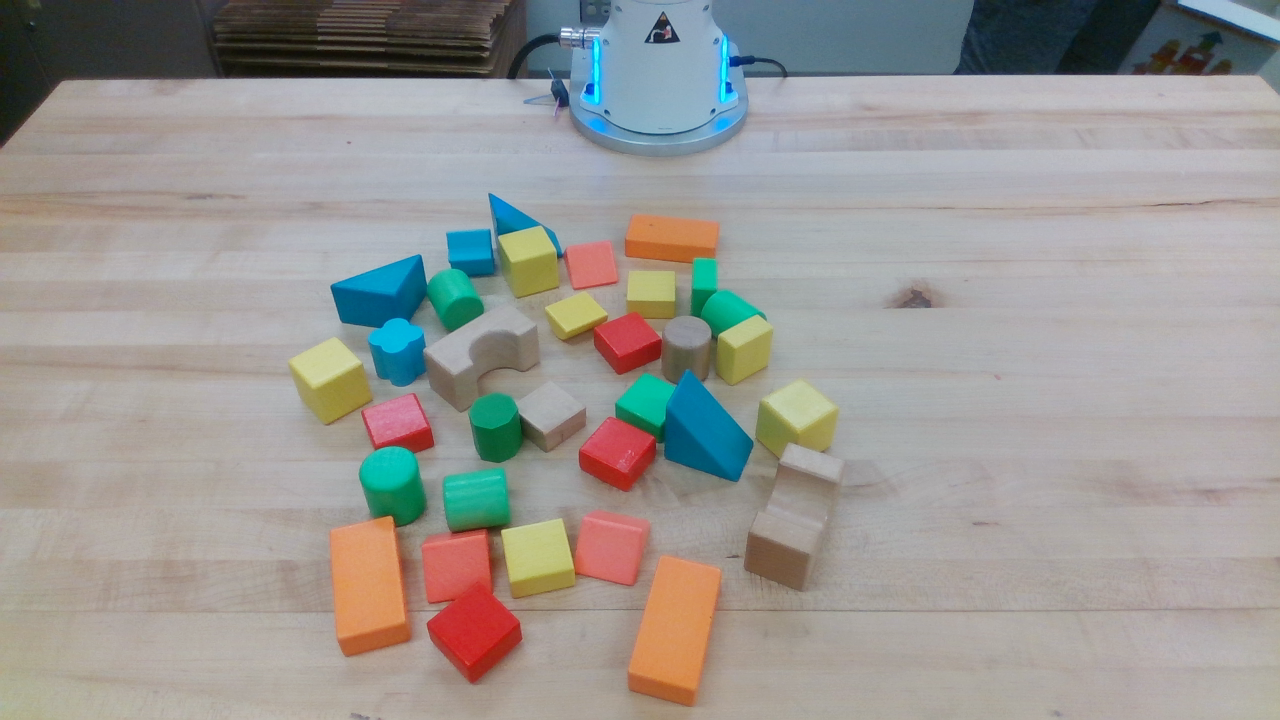}
\caption{Block-Medium}
\end{subfigure}
\begin{subfigure}{0.154\textwidth}
\includegraphics[width=\textwidth]{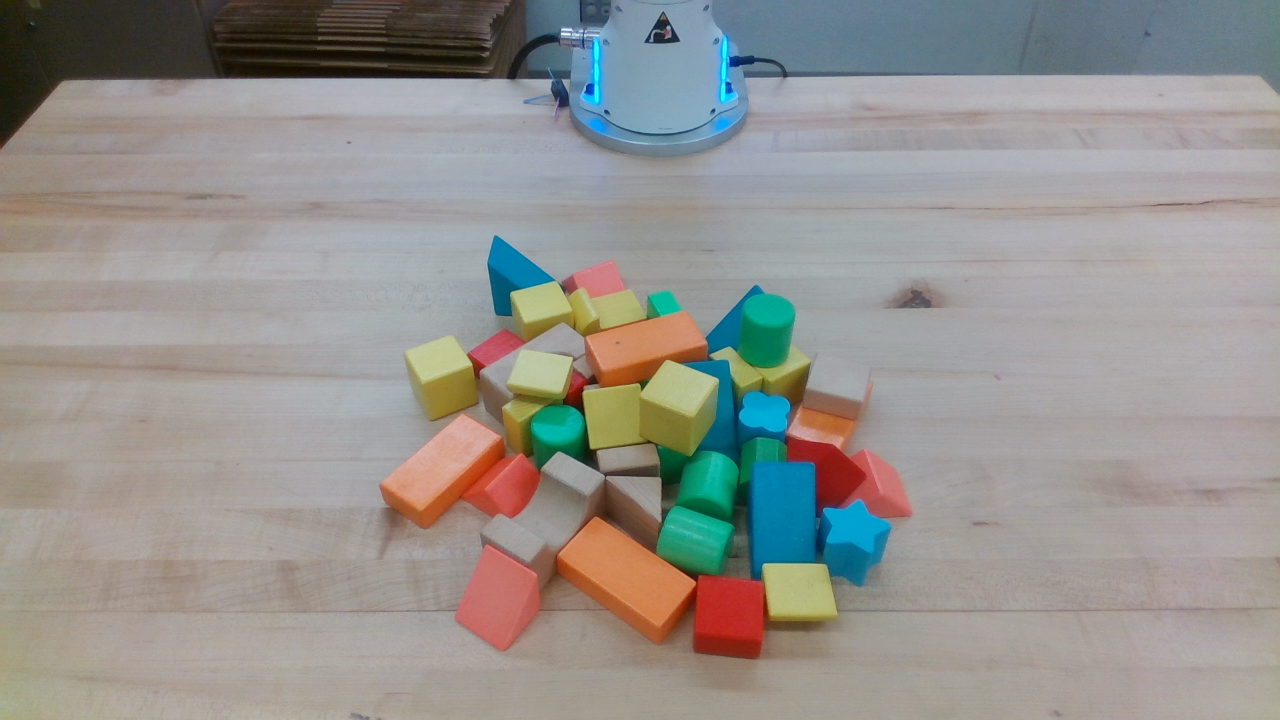}
\caption{Block-Dense}
\end{subfigure}
\caption{\textbf{Block Clutter in the real world.} As scene grows more complex from (a) to (c), knowledge of object relations significantly facilitates grasping activities.}
\label{fig:real_block}
\vspace{-8pt}
\end{figure}

\begin{figure}[t]
\begin{subfigure}{0.154\textwidth}
\includegraphics[width=\textwidth]{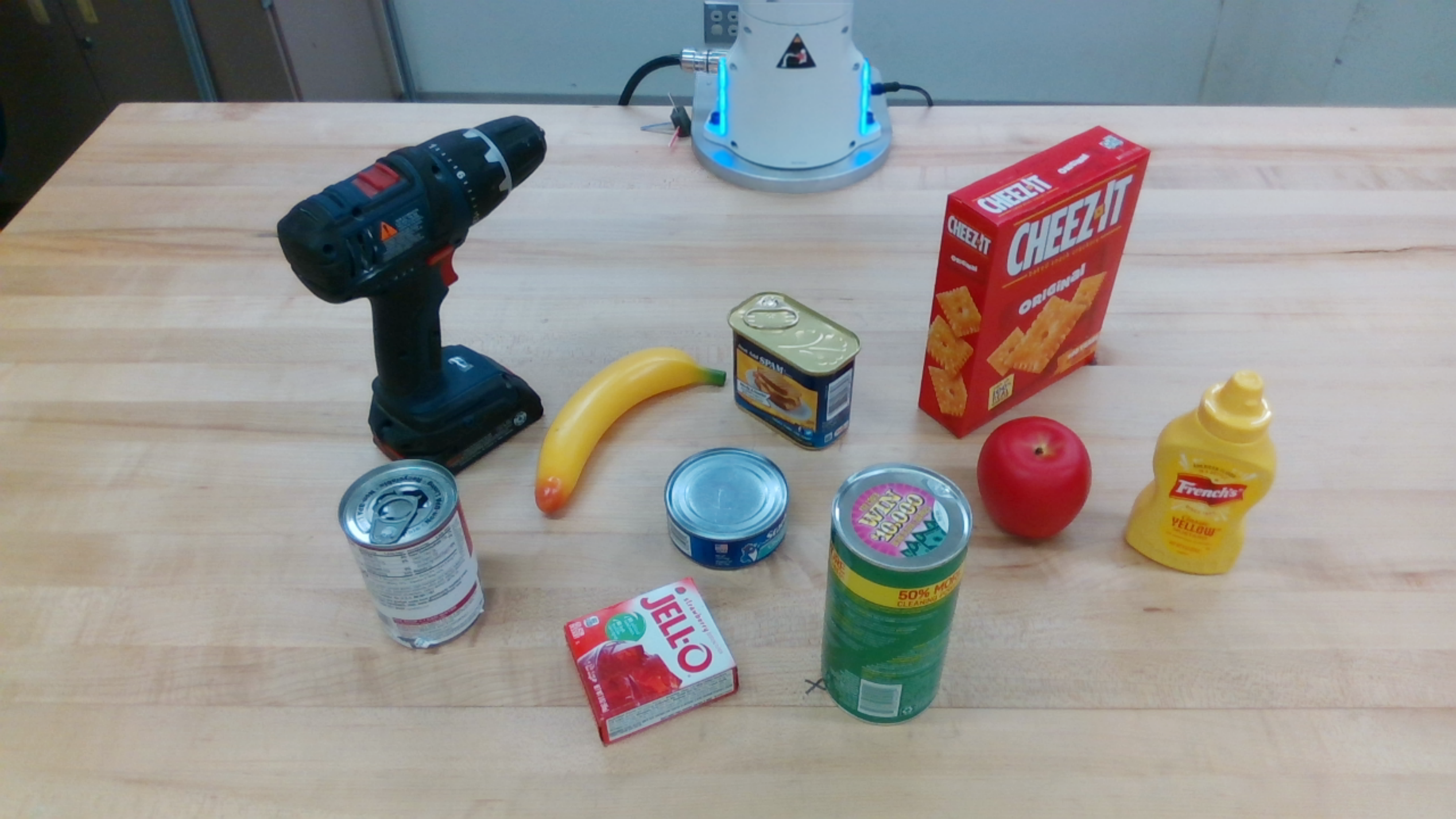}
\caption{Novel-Single}
\end{subfigure}
\begin{subfigure}{0.154\textwidth}
\includegraphics[width=\textwidth]{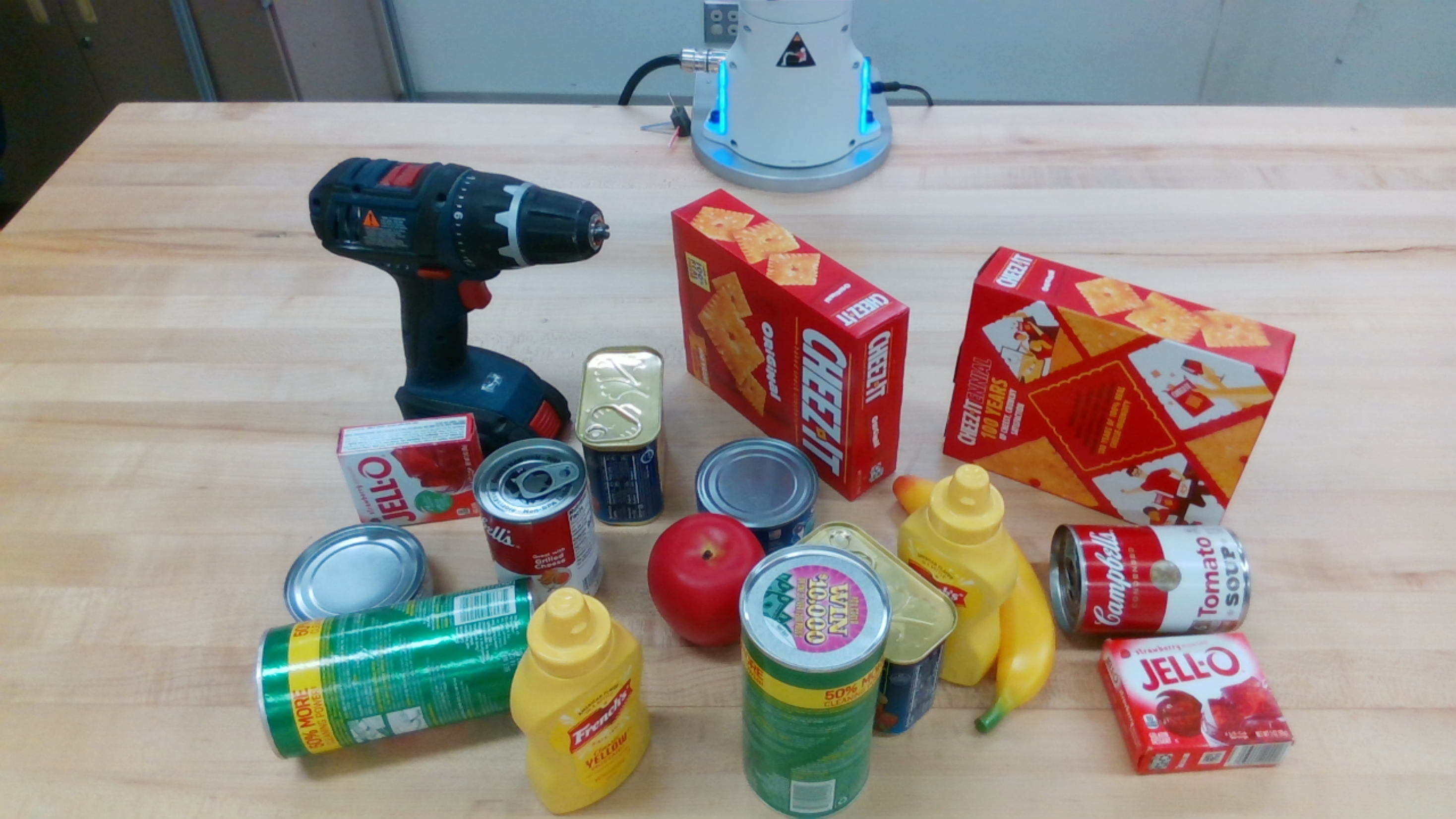}
\caption{Novel-Medium}
\end{subfigure}
\begin{subfigure}{0.154\textwidth}
\includegraphics[width=\textwidth]{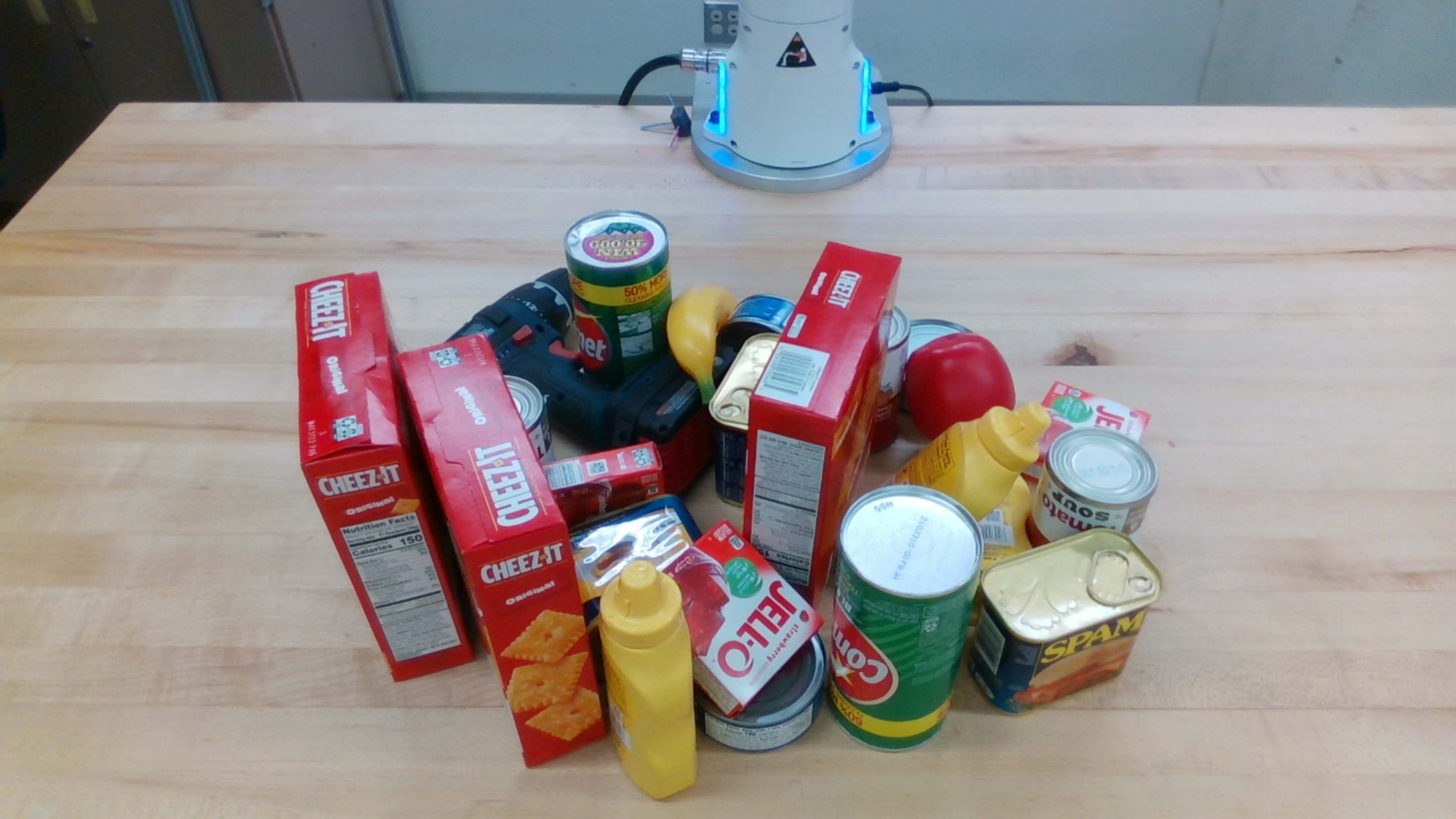}
\caption{Novel-Dense}
\end{subfigure}
\caption{\textbf{Novel scenes in the real world.} the scenes only include unseen objects and are arrange with increasing density.}
\label{fig:real_novel}
\vspace{-8pt}
\end{figure}

% \begin{figure}[t]
% \label{fig:simres_block}
% \centering
% \includegraphics[width=0.49\textwidth]{images/grasp_acc_realblock.pdf}
% \includegraphics[width=0.49\textwidth]{images/grasp_acc_realnovel.pdf}
% \caption{\textbf{Experiment Results in the Real World.} G2N2 achieved an overall 84.29\% and 81.48\% grasping accuracy in Block and Novel tests, respectively.}
% \end{figure}

\begin{table}[t]
\caption{Real-Robot Results in Block Clutter}
\label{tab:real_block}
\begin{center}
\begin{tabular}{|c||c||c||c||c||c|}
\hline
  & GSP & 6GN & GSP+AS & Ours\\
\hline
Block-Normal & 83.33 & 77.78 & 83.33 & \textbf{88.89}\\
\hline
Block-Medium & 77.78 & 72.22 & \textbf{83.33} & \textbf{83.33}\\
\hline
Block-Dense & 66.67 & 61.11 & 72.22 & \textbf{80.65}\\
\hline
\end{tabular}
% \vspace{-8pt}
\end{center}
\end{table}

\begin{table}[t]
\caption{Real-Robot Results in Novel Scenes}
\label{tab:real_novel}
\begin{center}
\begin{tabular}{|c||c||c||c||c||c|}
\hline
  & GSP & 6GN & GSP+AS & Ours\\
\hline
Novel-Single & 72.22 & \textbf{83.33} & 72.22 & \textbf{83.33}\\
\hline
Novel-Medium & 44.44 & 77.78 & 66.67 & \textbf{83.33}\\
\hline
Novel-Dense & 38.89 & 61.11 & 66.67 & \textbf{77.78}\\
\hline
\end{tabular}
\vspace{-8pt}
\end{center}
\end{table}

We experiment in the real world to further examine the effectiveness of the proposed grasping system. The setup consists of a Franka Emika Panda robot arm and an Intel RealSense D415 RGB-D camera, externally mounted on a tripod facing the robot. A single-view RGB-D image is taken as the input for all approaches. Given multiple instances of a target object, the goal is to retrieve one of them from the scene. For each round of testing, we randomly initialize three different scenes as shown in Fig.~\ref{fig:real_block}. Six target objects are grasped sequentially per scene for 3 rounds, for a total of 18 grasps per method. The first testing scenario includes wooden blocks similar to our training objects in simulation. The experimental results are summarized in Table~\ref{tab:real_block}. The performance of baseline approaches drop significantly with increasing object density, necessitating knowledge of object relations. Our approach achieves the highest grasping accuracy in all three scenarios, suggesting that it is the most efficient and sim-to-real transferable.

Following the same procedure, we test the generalizability to unseen objects with the arrangements shown in Fig.~\ref{fig:real_novel}. Note that all testing objects are distinct from the training ones in size, color, and shape, largely increasing the inference challenge. Table~\ref{tab:real_novel} compiles the results of the tests. In \textit{Novel-Single}, GSP+AS is equivalent to GSP since only one instance of the target object is presented. Their narrow focus struggles with larger objects. Although 6GN is trained with YCB objects and able to manage to excel in \textit{Novel-Single}, it only achieves a 61.11\% grasping accuracy in \textit{Novel-Dense} since it excessively grasp the more cluttered object. By analytically finding the less cluttered target, GSP+AS improves the grasping accuracy of GSP in \textit{Novel-Medium} and \textit{Novel-Dense} scenarios by more than 70\%. However, it heavily relies on collision detection, which is vulnerable to occlusions. Overall, the results are consistent with the simulation counterparts. With the help of G2N2, we reason about the spatial relations between objects; our 6-DoF grasping system achieves a 77.78\% grasping accuracy in \textit{Novel-Dense}, outperforming the other baselines by large margins.

\section{CONCLUSION}

We presented a 6-DoF grasping system that addresses a complex target-driven task, grasping one of the identical objects from a dense clutter. Our approach used the Grasp Graph Neural Network (G2N2) to harness spatial relations between objects. Additionally, we designed a more efficient shape completion-assisted 6-DoF grasp pose sampling method that facilitates both training and grasping. Throughout exhaustive evaluation in both simulated and real settings, the proposed grasping system achieved a 88.89\% grasping accuracy on scenarios with densely cluttered blocks and 77.78\% grasping accuracy on scenarios with densely cluttered novel objects.

%%%%%%%%%%%%%%%%%%%%%%%%%%%%%%%%%%%%%%%%%%%%%%%%%%%%%%%%%%%%%%%%%%%%%%%%%%%%%%%%
\bibliographystyle{IEEEtran}
\bibliography{IEEEabrv,IEEEexample}
\end{document}